\pdfoutput=1
\documentclass[11pt]{article}

\usepackage[]{acl}
\usepackage{tabularx}

\usepackage{times}
\usepackage{latexsym}
\usepackage{subfig}
\usepackage{multicol}
\usepackage{multirow}
\usepackage{graphicx}
\usepackage{amsmath}

\usepackage[T1]{fontenc}

\usepackage[utf8]{inputenc}

\usepackage{microtype}

\usepackage{inconsolata}

\usepackage{enumitem}
\usepackage{hyperref}

\title{Connecting the Dots: Evaluating Abstract Reasoning Capabilities of LLMs Using the New York Times Connections Word Game}

\author{Prisha Samadarshi\textsuperscript{1}~~~~~Mariam Mustafa\textsuperscript{1}~~~~~Anushka Kulkarni\textsuperscript{1}~~~~~Raven Rothkopf\textsuperscript{1}\\~~~~~\textbf{Tuhin Chakrabarty}\textsuperscript{2}\thanks{Equal contribution. \dag denotes work done at Columbia}\textsuperscript{\dag}~~~~~\textbf{Smaranda Muresan}\textsuperscript{1*} \\
 \textsuperscript{1} Department of Computer Science; Barnard College, Columbia University\\
 \textsuperscript{2} Department of Computer Science, Stony Brook University\\
 {\tt\small \{ps3203, mm5970, ajk2256, rgr2124\}@alum.barnard.edu, tuhin.chakrabarty@stonybrook.edu, smuresan@barnard.edu} \\
}

\begin{document}
\maketitle
\begin{abstract}

The \emph{New York Times} Connections game has emerged as a popular and challenging pursuit for word puzzle enthusiasts. We collect 438 Connections games to evaluate the performance of state-of-the-art large language models (LLMs) against expert and novice human players. Our results show that even the best-performing LLM, Claude 3.5 Sonnet, which has otherwise shown impressive reasoning abilities on a wide variety of benchmarks, can only fully solve 18\% of the games. Novice and expert players perform better than Claude 3.5 Sonnet, with expert human players significantly outperforming it. We create a taxonomy of the knowledge types required to successfully cluster and categorize words in the Connections game. We find that while LLMs perform relatively well 
 on categorizing words based on semantic relations they struggle with other types of knowledge such as Encyclopedic Knowledge, Multiword Expressions or knowledge that combines both Word Form and Meaning. Our results establish the \emph{New York Times} Connections game as a challenging benchmark for evaluating abstract reasoning capabilities in 
 AI systems.
\end{abstract}

\section{Introduction}

Abstract reasoning represents a person's ability to solve problems, identify patterns, and work with logical systems \citep{barrett2018measuring, johnson2021fast,ji2022abstract}. While the performance of large language models (LLMs) on arithmetic and language-based commonsense reasoning benchmarks has been the subject of recent analyses, it is unclear whether these LLMs possess abstract reasoning capabilities that are often challenging even for humans \citep{xu2023llms}. We propose the \emph{NYT} Connections Game as a test bed for investigating the abstract reasoning capabilities of both humans and large language models (LLMs).

\begin{figure}[!t]
    \centering
    \subfloat[The unsolved connections game presented to a player]{\includegraphics[scale=0.23]{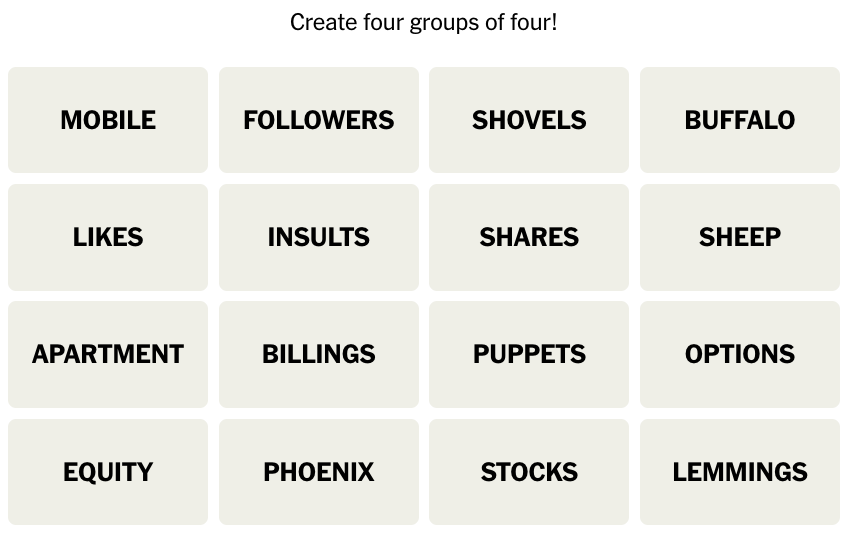}}\\
    \subfloat[The solved connections game with correct categories shown in ascending level of difficulty---straightforward (yellow) to tricky (purple)]{\includegraphics[scale=0.27]{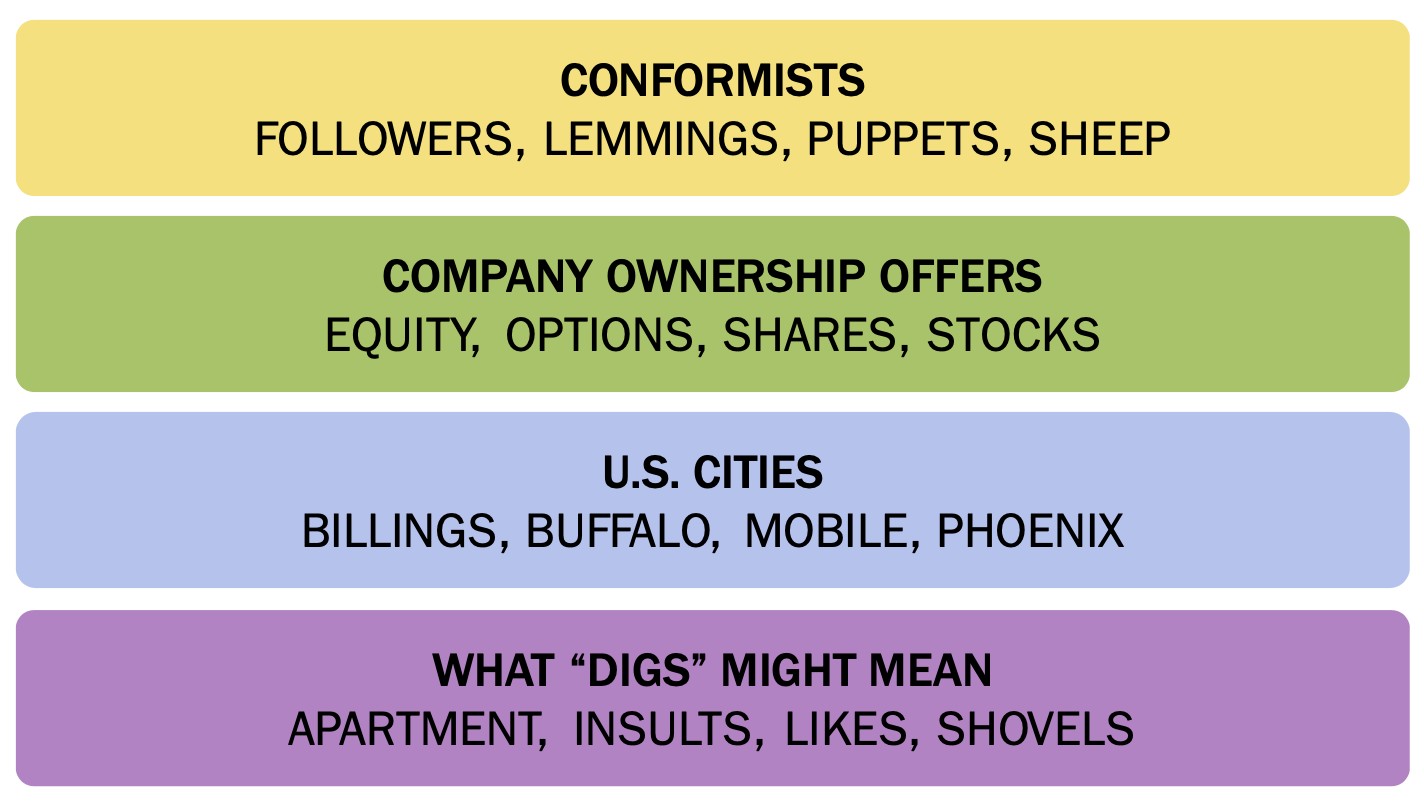}}
    \caption{Example from a \emph{NYT} Connections game}
    \label{fig:introFig}
\end{figure}

Connections is an engaging game launched by the \emph{New York Times} (\emph{NYT}) in June 2023. This daily game presents players with a 4x4 grid containing 16 words and tasks them with identifying four distinct clusters that link the corresponding four words in each cluster through some shared characteristics (Figures \ref{fig:introFig} [a] and [b]). 
Categories 1 (yellow), 2 (green), 3 (blue), and 4 (purple) are arranged in ascending level of difficulty. Category 1 is the most intuitive, while Category 4 is the hardest. For instance, in Figure \ref{fig:introFig} (b), the most straightforward category is "Conformists" \emph{\{Followers, Lemmings, Puppets, Sheep\}}, while the most challenging category includes \emph{\{Apartment, Insults, Likes, Shovels\}} and requires the understanding that a single word (in this case, "digs") can have multiple meanings that differ in etymology or sense, depending on the context.

While the task might seem easy, many words can be grouped easily into multiple categories, acting as red herrings. For instance, from the game in Figure~\ref{fig:introFig}, \emph{Likes, Followers, Shares, Insult} might be categorized as ``Social Media Interactions" at first glance. Unlike common categories (e.g., ``Fruit," ``Furniture”), the game is designed to promote ad hoc category formations that violate the correlational structure of the environment and are not well established in memory \citep{Barsalou1983AdHC}. To group words across proper categories, as shown in Figure \ref{fig:introFig} (b), a player must reason with various forms of knowledge spanning from \emph{Semantic Knowledge} (Conformists) to \emph{Encyclopedic Knowledge} (U.S. cities).

We test the capabilities of five state-of-the-art large language models, namely Google's Gemini 1.5 Pro \citep{team2023gemini}, Anthropic's Claude 3.5 Sonnet \citep{claude3}, OpenAI's GPT-4Omni \citep{gpt4o}, Meta's Llama 3.1 405B, \citep{llama3modelcard} and Mistral Large 2 \citep{mistral} on 438 distinct \emph{NYT} Connections games and compare them with human performance on a subset of these games.
Our experimental results show that while all LLMs can partially solve some games, their performance is far from ideal.
Even the best-performing model Claude 3.5 Sonnet (with few-shot and chain-of-thought prompting), can only solve 18\% of the games completely. In addition, we recruit human players at novice and expert levels of proficiency and compare their performance to Claude 3.5 Sonnet. Our results show that the \emph{NYT} Connections game serves as a challenging benchmark for reasoning, with novice players performing marginally better than Claude 3.5 Sonnet and expert players performing significantly better than Claude 3.5 Sonnet in solving games perfectly (Section \ref{sec:results}). 

In addition, to better understand the LLMs abstract reasoning capabilities or the lack thereof, we propose a taxonomy of knowledge required to group words into their respective categories (Section \ref{sec:reasoning}). Our analysis at Section \ref{sec:reasoning_results} shows that while LLMs are relatively better at reasoning that involve Semantic Relations, they struggle with other types of knowledge such as Multiword Expressions and combined knowledge about Word Form and Word Meaning (Section). 

Our code and data will be made available to the public at \url{https://github.com/mustafamariam/LLM-Connections-Solver}.

\section{Related Work}
The growing popularity of Large Language Models has led to an exciting array of research using natural language processing techniques for text-based games. Recent work has studied whether these models can act as players in agentic environments \cite{huang2024far,wang2023voyager, wu2023smartplay,noever2020chess} or conversational settings \cite{qiao2023gameeval}. In addition to acting as players, researchers have also tested the abilities of transformer-based language models in generating games \cite{ammanabrolu2021modeling,todd2023level,sudhakaran2024mariogpt,hu2024generating,chen2023gamegpt,merino2024making}

Recent research has explored applying large language models (LLMs) and other natural language processing (NLP) techniques to solve and generate text-based puzzles. \citet{wallace-etal-2022-automated} propose automatic ways of solving crossword puzzles by generating answer candidates for each crossword clue using neural question answering models and combining loopy belief propagation with local search to find full puzzle solutions. \citet{zhao2023solving} release PUZZLEQA, a multiple-choice dataset comprising 15 years of on-air Sunday Puzzle word-games and show that while ChatGPT can solve these questions with an accuracy of around 50\%, they still struggle with generating novel and engaging puzzles. Unlike the \emph{NYT} Connections game, PUZZLEQA relies on character-level word transformations compared to encyclopedic, associative, or semantic knowledge. \citet{rozner2021decrypting} examined the potential of using "cryptic crossword" clues as an NLP benchmark. 

Most relevant to our paper is the contemporaneous work by \citet{todd2024missed} who test the performance of various LLMs (from BERT and RoBERTA to GPT-4 and GPT-4 with Chain-of-Thought prompting) in solving the \emph{NYT}  Connections game. Our work is similar in that it utilizes the \emph{NYT} Connections puzzles as a means to investigate the abstract reasoning capabilities of state-of-the-art LLMs. However, our contribution focuses not only on the ability of LLMs to solve the game, but also on studying the types of knowledge required to so. Moreover, we evaluate both state--of-the-art models (open and closed weights) and humans (novices and experts). We also benchmark on a larger number of \emph{NYT} Connections game.
Additionally, \citet{todd2024missed}'s experimental setup mirrors the way the original \emph{NYT} Connections game is played (i.e., one category at a time, with an allotment of 4 incorrect guesses), while we require both humans and LLMs to provide all categories at once in only one attempt.

\begin{figure*}[!ht]
    \small
    \centering
    \includegraphics[scale=0.55]{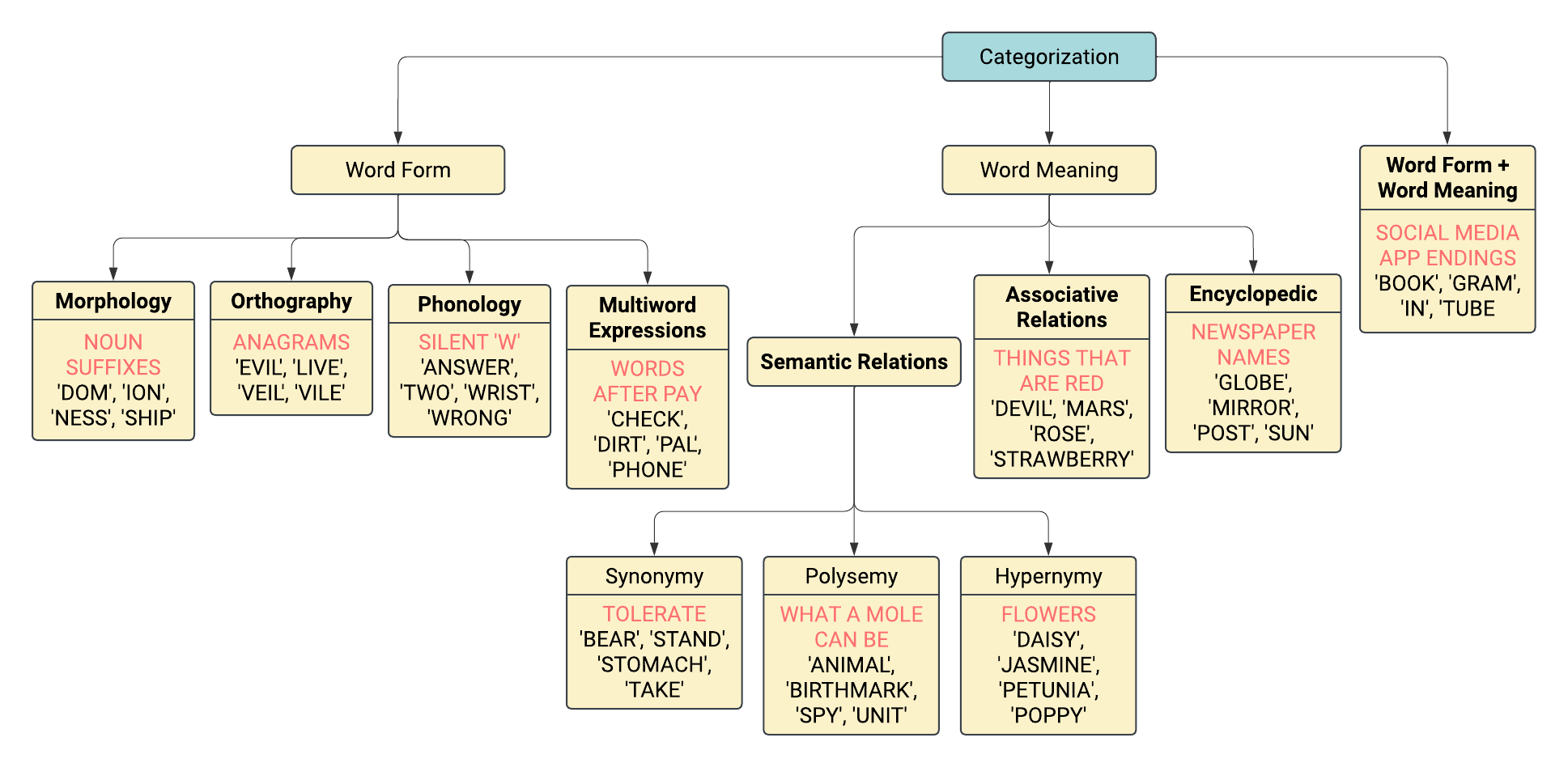}
    \caption{\label{knowledgetype}Proposed taxonomy of knowledge types required to solve the Connection games. In our evaluation we used the categories in bold}
\end{figure*}

The word association task \cite{10.1093/brain/2.2.149} has been used extensively in psychological and linguistic research as a way of measuring connections between words in the mental lexicon. Responses in word association tasks have informed what we know about the structure and organization of semantic memory and the mental lexicon \cite{de2008word}. In this work, we similarly show how one must utilize semantic and associative memories to solve the \emph{NYT} Connections game.

\citet{chollet2019measure} proposed the Abstraction and Reasoning Corpus (ARC), built upon an explicit set of priors designed to be as close as possible to innate human priors and argued that it can be used to measure a human-like form of general fluid intelligence, enabling fair general intelligence comparisons between AI systems and humans. Recently \citet{xu2023llms} show that GPT-4 solves only 13/50 of the most straightforward ARC tasks, demonstrating a significant gap in the abstract reasoning capabilities of LLMs. Prior work has also studied abstract reasoning in Neural Networks \cite{barrett2018measuring} in the presence of distracting features \cite{zheng2019abstract}. Our work builds upon these and presents the \emph{NYT} Connections game as a compelling benchmark for abstract reasoning capabilities of LLMs in the presence of distractors.

\section{Data} 
\subsection{Collection}\label{sec:data}
To gather the necessary data, we found an archival site consisting of all possible answer choices and their corresponding categorizations. As the \emph{NYT} does not maintain an archive of \emph{NYT} Connections puzzles, we resorted to an external, third-party site for data collection.\footnote{\url{https://tryhardguides.com/nyt-connections-answers/}} Our data spans daily problems from the conception of \emph{NYT} Connections in June 2023 to August 2024. In total, we gather 441 distinct games, out of which 3 are used for few-shot prompting, while the remaining 438 comprise the dedicated test set.

\subsection{Types of Reasoning}\label{sec:reasoning}

Investigating the relationship between words offers insights into both the structure of language and the influence of cognition on linguistic tasks \citep{stella2018multiplex}. To solve the \emph{NYT} Connections game, players must draw on certain aspects of word knowledge, such as a word’s meaning or form and sometimes both simultaneously. To deepen our understanding, we bucket each $<$category, grouping$>$ into the types of knowledge that are primarily required to solve them. Three linguists annotate a total of 1,752 samples coming from 438 games into 3 broader categories which give rise to 8 subcategories (Figure \ref{knowledgetype}). We take majority voting to arrive at a unique label for each $<$category, grouping$>$. We restrict annotations to sub-categories and not sub-sub-categories (e.g., Types of Semantic Relations). We obtain a Fleiss Kappa of 0.78 showing substantial agreement.

\subsubsection{Word Form}
Word Form refers to the specific shape or appearance a word takes in a given context. It encompasses various aspects of a word's structure and representation and is broadly used in the \emph{NYT} Connections game, testing knowledge on \textit{Phonology, Orthography, Morphology} and \textit{Multiword Expressions}. Phonology in Word Form deals with the sound structure of words, including pronunciation, stress patterns, and phonetic variations. Orthography relates to the conventional spelling system of a language, which may not always directly correspond to pronunciation. Morphology examines the internal structure of words, including roots and affixes and how they combine to create meaning. For example, as shown in Figure \ref{knowledgetype}, one needs morphological knowledge to group \emph{Dom, Ion, Ness,} and \emph{Ship} as ``Noun Suffixes." Similarly, one needs to rely on phonological knowledge of the sound patterns of \emph{Answer, Two, Wrist,} and \emph{Wrong} to categorize them as ``Silent `W'." Multiword Expressions (MWE) have a fixed or semi-fixed form and are typically non-compositional (i.e., their meaning cannot be predicted from their individual components) \cite{moon1998fixed}.

\subsubsection{Word Meaning}
\paragraph{Semantic Relations: }The majority of instances in the \emph{NYT} Connections game require possessing knowledge of \emph{semantic relations} \cite{murphy2003semantic, cruse1986lexical}, such as synonymy (words with the same meaning), hypernymy/hyponymy (relation between a generic term and its specific instance), and homonymy and polysemy (many possible meanings of a word). Figure \ref{knowledgetype} shows three examples of groups that use such semantic relations.

\paragraph{Associative Relations}
Prior work has studied models of automatic priming for word identification that are typically divided into groups based on associative relations (e.g., spreading activation) and others based on semantic similarity (e.g., distributed models) \cite{thompson1998effects}. In contrast to category members that share semantic features and category nodes, Associative Relations are elements of specific situations or thematic contexts, with little or no overlap between their semantic features (e.g. ``Things that are red": \emph{Mars} and \emph{Strawberry}; Figure \ref{knowledgetype}) \cite{rose2016cumulative, shanks1995psychology, Barsalou1983AdHC}.

\paragraph{Encyclopedic}
We notice that to group certain sets of words into their proper categories, one needs knowledge that spans beyond concepts and relies on entities in the real world found in knowledge bases such as Wikipedia \cite{mihalcea2007wikify}. This can be seen in Figure \ref{knowledgetype}, where, to bucket the words \emph{Globe, Mirror, Post,} and \emph{Sun} into the category of ``Newspaper Names," one needs to possess knowledge that Globe refers to the \emph{Boston Globe}, Mirror to the \emph{Daily Mirror}, a UK tabloid, Post to the \emph{Washington Post} and Sun to \emph{The Sun}, another UK tabloid. We label this type of knowledge Encyclopedic. 

\subsubsection {Word Form + Word Meaning} Some of the hardest examples in the \emph{NYT} Connections game require reasoning of both word form and meaning. For instance, the example in Figure \ref{knowledgetype} shows that to group the words \emph{Book, Gram, In,} and \emph{Tube}, one needs to identify that they are essentially parts of closed compounds (Face+Book, Insta+Gram, Linked+In, You+Tube) that also represent popular social media apps. This categorization requires the use of knowledge on Word Meaning (Encyclopedic) as well as Word Form (Morphology).

\begin{table*}[!ht]
\centering
\small
\begin{tabular}{|cc|ccc|c|}
\hline
\multicolumn{2}{|c|}{Word Form}                                                                                                                                         & \multicolumn{3}{c|}{Word Meaning}                                                                                                                                                     & \begin{tabular}[c]{@{}l@{}}Word Meaning\\ + Word Form\end{tabular} \\ \hline
\multicolumn{1}{|l|}{\begin{tabular}[c]{@{}l@{}}Phonology/Orthography/\\ Morphology\end{tabular}} & \begin{tabular}[c]{@{}l@{}}Multiword\\ Expressions\end{tabular} & \multicolumn{1}{l|}{\begin{tabular}[c]{@{}l@{}}Semantic\\ Relations\end{tabular}} & \multicolumn{1}{l|}{\begin{tabular}[c]{@{}l@{}}Associative\\ Relations\end{tabular}} & Encyclopedic & \multirow{2}{*}{92}                                                   \\ \cline{1-5}
\multicolumn{1}{|c|}{44}                                                                               & 168                                                             & \multicolumn{1}{l|}{1045}                                                         & \multicolumn{1}{l|}{137}                                                             & 266          &                                                                       \\ \hline
\end{tabular}
\caption{\label{knowledgedistr} Distribution of different knowledge types required to categorize words across 438 games}
\end{table*}

\section{Experimental Settings} \label{sec:experiment}
\subsection{LLMs as Game Players}
To test the capabilities of large language models in solving the \emph{NYT} Connections game, we rely on recent advancements in in-context learning and chain-of-thought prompting \cite{wei2022chain}. We provide 3 complete examples in our few-shot prompt along with rules and common strategies that players must use to solve the game. We also elicit chain-of-thought reasoning \cite{wei2022chain}, requiring models to explain their groupings and categories chosen. Formulation of the prompt involved trial and error; the first iteration of the prompt included the \emph{NYT} Connections game instructions provided by the \emph{New York Times} \citep{liu_connections_2023}, and included three demonstrations with gold labels asking the LLM to explain its reasoning in a step-by-step manner \cite{wei2022chain}. We ran this first prompt on a development set of 30 games, using 5 LLMs. After identifying commonalities in the types of errors made by the LLMs while playing the game, we added additional instructions, specified the response format, and included some tips from a \emph{NYT} article about playing the \emph{NYT} Connections game \citep{aronow_how_2023}. The entire prompt is in Appendix~\ref{appendix:prompt}. To ensure consistency and fairness in performance, we prompt 5 LLMs --- Gemini 1.5 Pro, Claude 3.5 Sonnet, GPT-4o, and Llama 3.1 405B and Mistral Large 2 --- with the same input. We use the default sampling parameters (temperature and top\_p) and the scoring schema outlined in Section~\ref{sec:scoring} to evaluate how all models perform in solving 438 \emph{NYT} Connections games spanning from June 2023 to August 2024.

\subsection{Humans as Game Players}
Alongside LLMs, we recruited 17 human evaluators in two subgroups: 12 novice players with little to no prior experience playing \emph{NYT} Connections and 5 expert or regular \emph{NYT} Connections players. The novice and the expert evaluators were peers of the first four authors,  who volunteered to participate without any payment. We designed a human evaluation interface and randomly sampled 100 games from our test set. Appendix~\ref{sec:humanevalinterface} has more information about the interface. The first screen displays an abridged version of the instructions from the LLM's prompt so as to not overwhelm the human players. To ensure that the humans solve the game in a manner comparable to the LLMs setup, they were given one try to solve the game (i.e., make all 4 categorizations at once). This aligns with \cite{todd2024missed}'s \textit{challenge mode}. 

Playing these games is a significant cognitive burden. As such, each novice human evaluator played around 8-12 distinct games for a total of 100 randomly sampled games out of the 438 in the test set, and expert participants each played 10 games for a total of 50 randomly sampled games.

\subsection{Evaluation Criteria} \label{sec:scoring}
Our scoring schema was developed as a means to numerically interpret the outcome of each \emph{NYT} Connections game and to standardize the comparison across LLMs and human players. We outline two processes to obtain \emph{clustering} and \emph{categorical reasoning} scores for a game of Connections.

\begin{figure*}[!ht]
    \small
    \centering
    \includegraphics[scale=0.5]{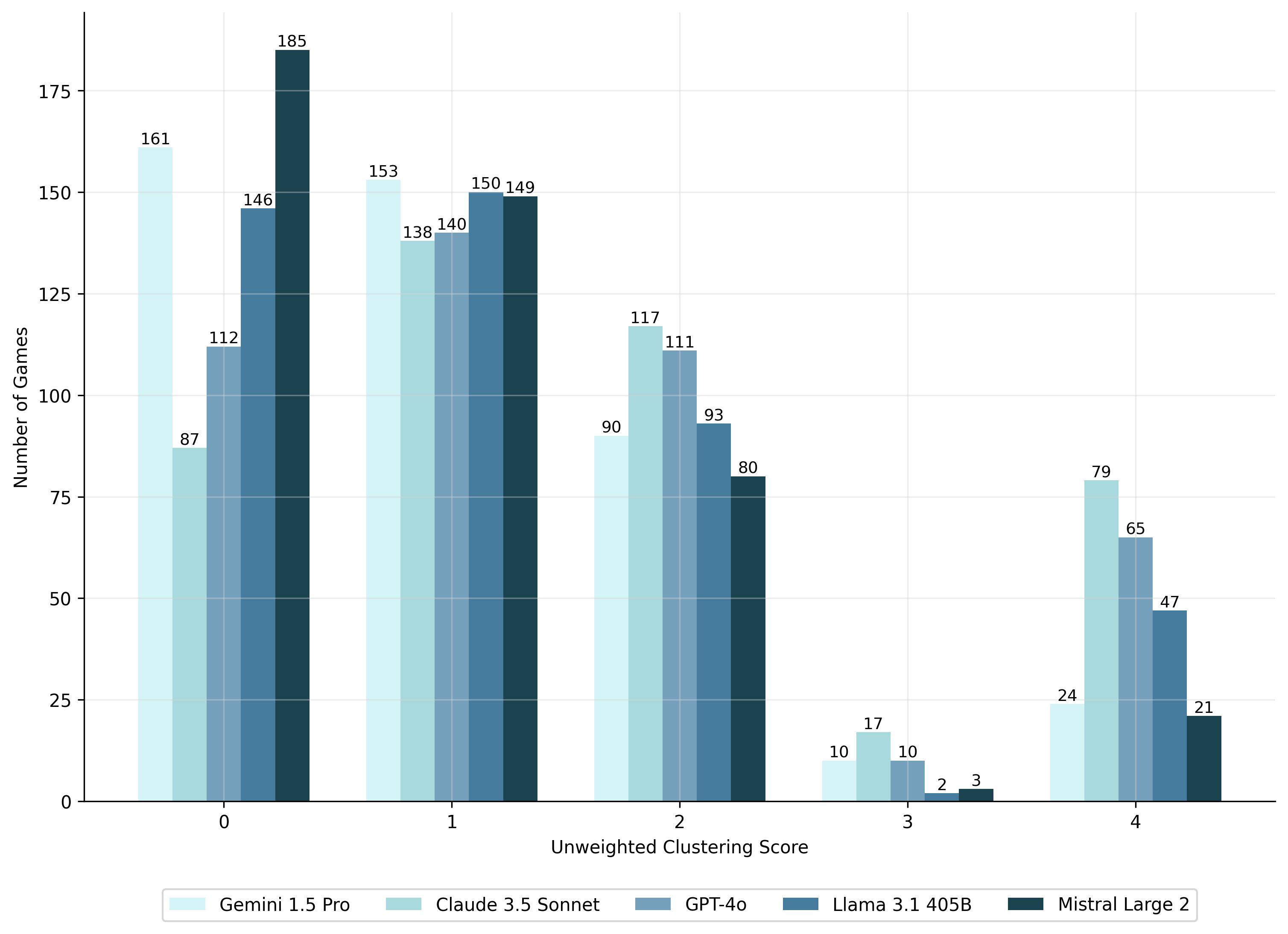}
    \caption{Frequency of unweighted clustering scores for 5 LLMs across 438 games. The number of games in which the respective unweighted clustering score was achieved is atop each bar. 0 means no correct clusters while 4 means all correct clusters}
    \label{fig:classication_score}
\end{figure*}

\subsubsection{Clustering Score} 
The clustering score evaluates the ability to correctly group all the words in the game. We consider two clustering scores. The first, or the \emph{unweighted clustering score}, is calculated independently of the categories' supposed difficulty. In this simple scoring mechanism, we allocate one point for each correct cluster (when all 4 words in the group classified by the player match the 4 words in the gold category/grouping). Ideally, a player's score should be close to the maximum of 4, signifying that all 4 groups were correctly identified. The equation is as follows:
\begin{equation}
score = n_0 + \ldots + n_3
\end{equation}
where $n_x=1$ for each correct grouping $x$ and $n_x=0$ for each incorrect grouping.

The second score, referred to as the \emph{weighted clustering score}, takes into account the difficulty of each grouping. The worst weighted clustering score a player can obtain is 0, meaning that no words were grouped correctly. Ideally, a player's score should be close to the maximum of 10, signifying that all 4 categories were correctly classified. The equation for this score is as follows:
\begin{equation}
score = n_0 \cdot w_0 + \ldots + n_3 \cdot w_3
\end{equation}
where $n_x$ represents one of the 4 categories and is always equal to 1 for each category $x$. The reward procedures are as follows: $w_0 = 1 $ for a Yellow (most straightforward) correct grouping, $w_1 = 2 $ for a Green correct grouping, $w_2 = 3 $ for a Blue correct grouping, $w_3 = 4 $ for a Purple (trickiest) correct grouping. Our schema for the clustering scores does not incorporate the number of tries as a variable, since in our setup the LLMs are prompted once and take one try to solve the game.

\subsubsection{Categorical Reasoning}
While the weighted and unweighted clustering scores are calculated for LLMs and humans, the \emph{categorical reasoning score} is used only for the LLMs' responses. If all 4 words in a category are correctly identified by an LLM, we conduct further analysis to evaluate whether the LLM reasoned correctly \emph{why} the words in the groups belong together. We make this distinction in our evaluation so that---in conjunction with the taxonomy of knowledge for \emph{NYT} Connections categories (Section~\ref{sec:reasoning})---we can assess the types of reasoning that the LLMs are most or least adept in. Since our prompt asks the LLM to include the category name and share the reasons why it grouped words, we can evaluate whether the reasoning in its response is semantically analogous to the gold \emph{NYT} Connections-provided category name. The decision of semantic equivalence between LLMs output and gold is done manually by a human judge to ensure accuracy. 

\section{Results}\label{sec:results}
\subsection{LLM performance}
Overall, we find that Claude 3.5 Sonnet performs best across thw 438 games. Figure ~\ref{fig:classication_score} shows the unweighted clustering scores for all 5 LLMs. Claude has the lowest percentage of games in which it made no correct clustering, at about 20\%, and most games solved perfectly at 18\%. Mistral Large 2 performs the worst overall. It could not make any correct clusters for 42\% of the games, and it perfectly solved the least amount of games. %SM I cannot translate easily from Fig 2 to these percetance points. you need to relate this text to Figure 3
\begin{figure}[!ht]
    \includegraphics[scale=0.35]
    {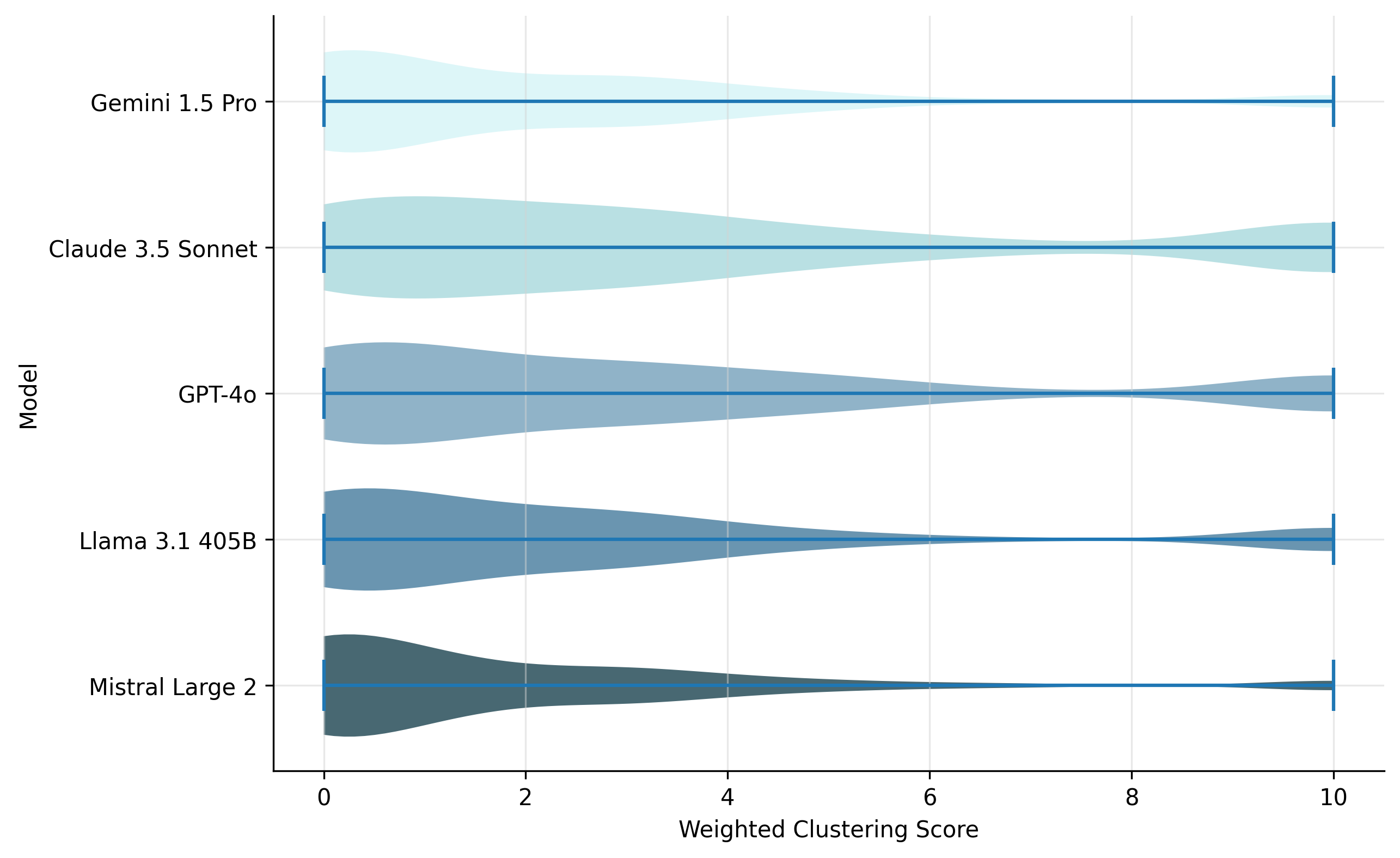}
    \caption{\label{fig:_weighted_score}Spread of weighted clustering scores for each model across 438 \emph{NYT} Connections games}
    %\vspace{-0.8em}
\end{figure}
In terms of weighted clustering scores for each model (Figure \ref{fig:_weighted_score}), Gemini 1.5 Pro and Mistral Large 2 show similar results. Most of their scores are concentrated before 2, showing that these models had a higher ability to correctly classify the easiest or second easiest categories. GPT-4o and Claude 3.5 Sonnet had most of their weighted clustering scores concentrated before 5, meaning they were better at classifying more and harder categories. Weighted clustering scores $\geq$ 8 are very rarely represented in all the models. Appendix~\ref{appendix:llm_performance} contains a more detailed breakdown. 

\subsection{Human Performance}
In human performance, we measure both novice and expert players against the best overall performing Claude 3.5 Sonnet. For the 100 games played by novices and 50 games played by experts, we compare the same games played by Claude 3.5 Sonnet.
\subsubsection{Novice Players}
\begin{figure}[!ht]
    \includegraphics[scale=0.47]{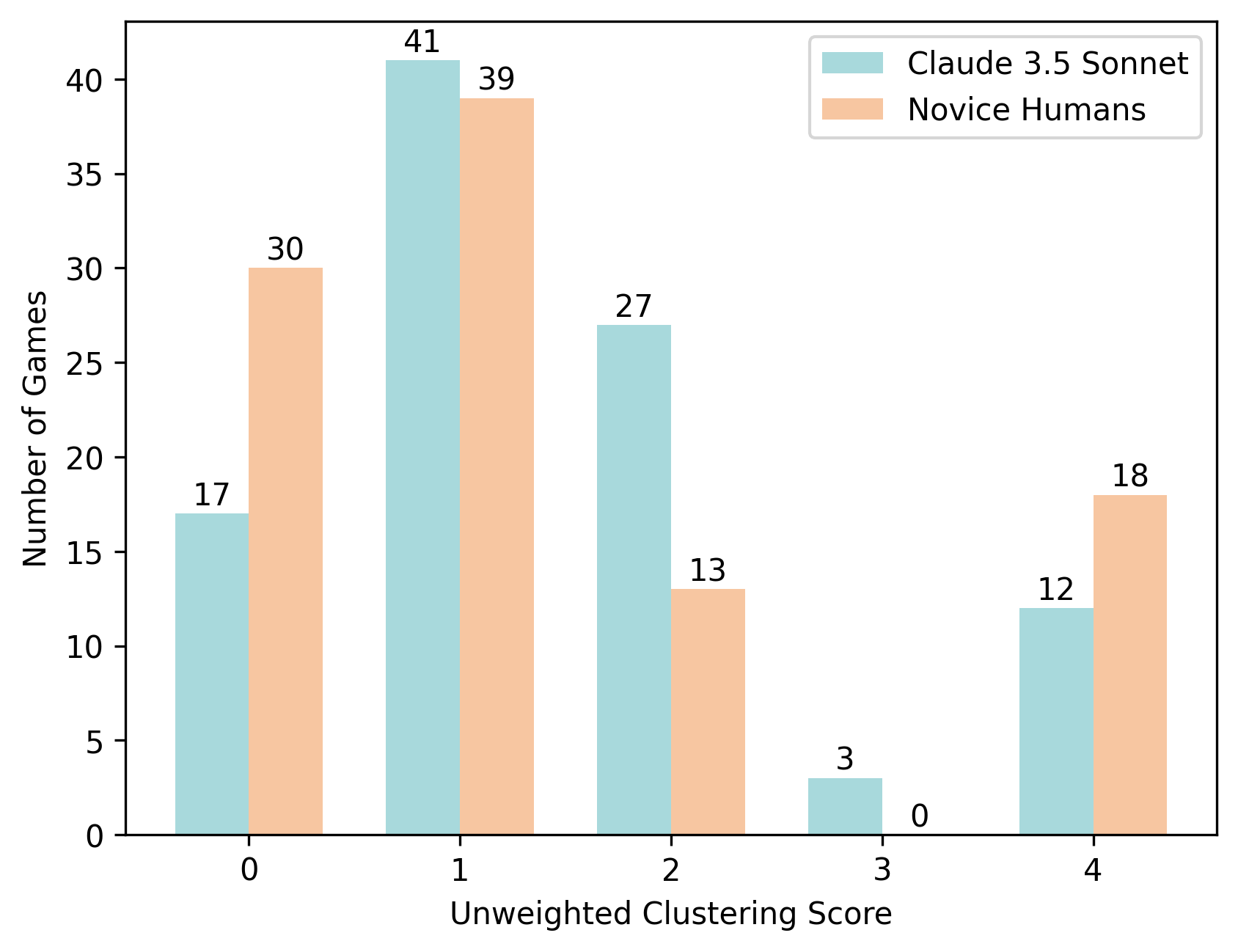}
    \caption{\label{fig:novice_classification}Frequency of clustering scores of Claude 3.5 Sonnet and 12 novice humans across 100 games}
\end{figure}
In the 100 games that the novice players completed, their average unweighted clustering score was 1.37, slightly worse than Claude's average of 1.52 in the same 100 games. Claude and novice humans also had similar weighted clustering score distributions. More details are in Appendix~\ref{appendix:human_performance}. Due to the setup of the human interface, humans could not receive a clustering score of 3 (if humans correctly solve 3 groupings, the fourth is also correct). Because of Claude's imperfect instruction-following abilities (repeating or omitting words), it was still able to obtain a clustering score of 3, as shown in Figure~\ref{fig:novice_classification}.

\subsubsection{Expert Players}
\begin{figure}[!ht]
    \includegraphics[scale=0.47]{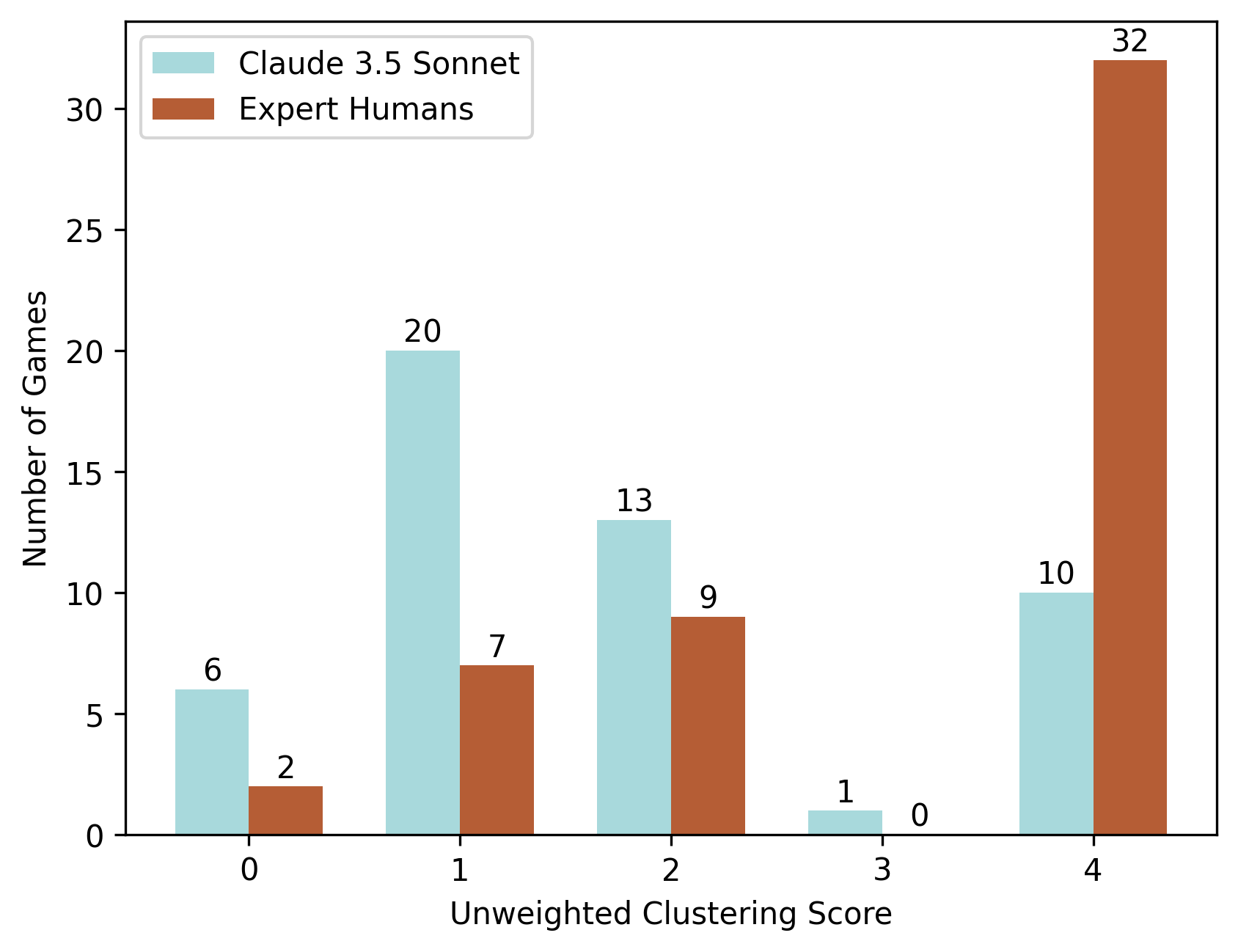}
    \caption{\label{fig:expert_classification} Frequency of clustering scores of Claude 3.5 Sonnet and 5 expert humans across 50 games}
\end{figure}
Expert human players performed significantly better than novices and Claude 3.5 Sonnet, with an average clustering score of 3.06 compared to Claude's 1.78 (on the same 50 games). The distribution of weighted clustering scores is also far more right-skewed (see Appendix~\ref{appendix:human_performance} for more). Figure~\ref{fig:expert_classification} shows that experts perfectly solved over 60\% of the 50 games, while Claude 3.5 Sonnet fully solved 20\% of these.

\section{Discussion}\label{sec:disc}

\begin{figure*}[!ht]
    \small
    \centering
    \includegraphics[scale=0.58]{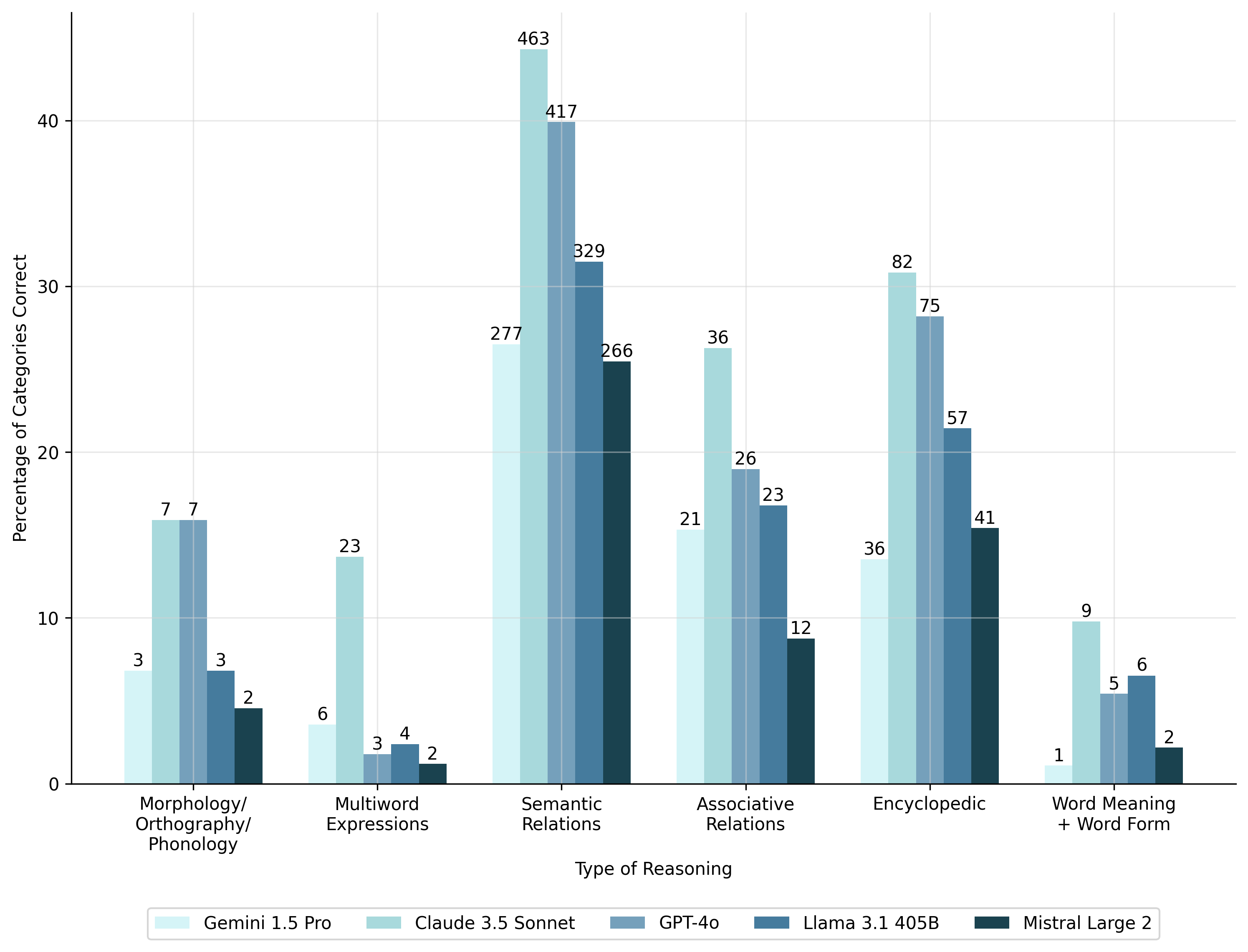}    \caption{\label{fig:llm_performance_reasoning} Percentage of categories from each knowledge type correctly classified and reasoned by the models across 438 games. The counts of categories correctly reasoned are displayed above each bar.}
\end{figure*}

\subsection{What type of reasoning is hardest for LLMs?} \label{sec:reasoning_results}

To answer this question we rely on our taxonomy of reasoning types introduced in Section \ref{sec:reasoning}. The breakdowns of the reasoning types for the 1752 categories in our 438-game dataset are shown in Table ~\ref{knowledgedistr}.  Figure~\ref{fig:llm_performance_reasoning} shows the performance of each LLM by reasoning type from our taxonomy of knowledge. Because the total counts of types of reasoning required across the 1,752 categories are unbalanced, the count of categories reasoned correctly is shown above each bar. The categories were only counted as correct if the model both clustered all the words correctly and justified its reasoning in a manner semantically analogous to the gold \emph{NYT} Connections-provided category name.    

The patterns in performance across the types of reasoning parallel the LLMs' overall performances for the most part, with Llama 3.1 405B's performance in Multiword Expressions and Word Meaning + Word Form  (slightly better than GPT-4o) and Mistral Large 2's performance in Word Meaning + Word Form (slightly better than Gemini 1.5 Pro) defying this pattern. The performance in reasoning categories across models is ranked from best to worst as follows: \textit{Semantic > Encyclopedic >  Associative > Morphology/Orthography/Phonology > Multiword > Word Meaning + Word Form}.  This aligns with the game makers' perceived level of category difficulty, as Word Meaning categories appear most often as yellow or green (easier) groupings, while Multiword Expressions and Word Meaning + Word Form categories are usually the purple groupings (most difficult). This is also consistent with the fact that the two types of reasoning LLMs performed best in---Semantic Relations and Encyclopedic Knowledge---exist in web-scraped information included in pre-training data, while other types of knowledge are more obscure and require iterative and deductive reasoning within each game. In every type of reasoning, however, LLMs perform with less than 50\% accuracy. Though they may be clustering words together accurately, they are not always justifying these clusters correctly, and this is visible in the differences between the frequencies of unweighted clustering scores of 4 and categorical reasoning scores of 4 (Appendix~\ref{appendix:llm_performance} Tables~\ref{table:llmclassification} and \ref{table:categorical}).

Since humans were not asked to provide justifications for their clusters, we do not include reasoning comparisons between LLMs and humans.

\subsection{How do distractors prevent LLMs and humans from correct categorization?}
The \emph{NYT} Connections game is often formulated with item overlap in mind, according to the \emph{NYT} Connections puzzle creator \citep{liu_how_2023}. These distractors, or red herrings, make the game far more challenging. Red herrings can appear in two ways---as a \emph{red herring category} or \emph{red herring word}. In the former case, 3 ultimately unconnected words seem to form a category of their own with 1 word missing. In the latter, a category seems applicable to more than 4 words, but the extras belong to a separate grouping. Examples of each of these types of red herrings are in Appendix~\ref{appendix:redherrings}.

Mistakes resulting from red herrings often occur in categories related to Associative Knowledge. Though the words may be associated in one dimension, the LLMs fail to conduct step-by-step reasoning to find another, perhaps more obscure, grouping (in the case of red herring categories) or the outlier (in the case of red herring words). 

\subsection{How often do LLMs group the words correctly but present incorrect reasons?}
To measure the disparity between LLMs making correct clustering and providing the correct reasoning or category name for their choice, we use a measure calculated from the clustering and categorical reasoning scores. Since the categorical reasoning score is the number of categories reasoned correctly and the clustering score considers whether the grouping was correct independent of the reason behind it, $\frac{\text{categorical reasoning}}{\text{unweighted clustering}}$ tells us how common it is for LLMs to cluster categories correctly by chance.
\begin{table}[!ht]
\centering
\begin{tabular}{lccc}
Model & Average Ratio \\
\hline
Gemini 1.5 Pro & 0.76 \\
Claude 3.5 Sonnet & 0.87 \\
GPT-4o & 0.84 \\
Llama 3.1 405B & 0.81 \\
Mistral Large 2 & 0.82\\
\end{tabular}
\caption{\label{table: ratio}Average categorical reasoning to unweighted clustering score ratio by model}
\end{table}
The average ratios in Table~\ref{table: ratio} are fairly high, close to or above 80\%. The highest overall performing models Claude 3.5 Sonnet and GPT-4o have the highest ratios as well. Though it is fairly uncommon that a model will correctly group words without correctly naming the reasoning behind that grouping, there are very few instances where models received both a clustering score of 4 (fully solved game) and a categorical reasoning score of 4. 

\subsection{How can future work improve on such a benchmark?}
Instead of choosing the first grouping, strategies grounded in System2 Thinking \cite{evans2003two} could improve performance on such a benchmark. Generating multiple chains-of-thought reasoning and learning a model that assigns a higher reward to the correct reasoning chain prevents the model from arriving greedily to a suboptimal categorization. Allowing LLMs to solve the game one category at a time and incorporating the feedback present to humans in the original \emph{NYT} Connections game, such as whether a grouping is correct (and what difficulty level it is by color), incorrect, or one word away from a correct grouping, may improve performance as well. Retrieval Augmentation from WordNet or dictionaries for lexical connotations \cite{allaway2020unified} could further improve such categorization. Finally, creating synthetic training data and training an LLM on this task could further close the gap between expert human and LLM performance. We leave such exploration for future work.

\section{Conclusions}\label{sec:conclusion}
We introduced the \emph{NYT} Connections games as a benchmark to test abstract reasoning in state-of-the-art LLMs and evaluate their performance against expert and novice human players. We find that Claude 3.5 Sonnet performs best, although it is still no match for expert human players. By examining the performance through our knowledge taxonomy, we obtain a more solid understanding of areas in which LLMs can improve.
Although most LLMs possess Word Meaning reasoning capabilities, they struggle with Multiword Expressions and combined knowledge categories. In addition,  red herrings pose a challenge to current LLMs.
Overall, we find that solving the \emph{NYT} Connections requires 
a breadth of different knowledge types, which current LLMs do not seem to fully master.

\section{Limitations}
Many of the limitations in this section stem from the lack of data available for\emph{NYT} Connections games and disparities in the comparison between LLMs and humans. Because it is a fairly recent invention and only one puzzle is released per day, there are only a few hundred games available. Since there are some category patterns learned through frequent play, ideally, a model trained on past \emph{NYT} Connections games might bridge the gap between LLM and expert human evaluators' performance. 

We acknowledge that human evaluators were not required to add justifications for the groupings they made. This might have made performance comparisons between humans and LLMs for the types of reasoning more equal. Additionally, because a score of 3 was impossible in the human evaluation interface, we cannot be certain that humans were adept in the type of knowledge of their last category grouped, as this could simply have been a matter of grouping all options left. Other limitations of human evaluators include that because they were all peers or acquaintances of the paper's authors, sampling bias could exist. Though the age range of the humans recruited was 14-60, other demographic factors that not have been accounted for in this sample. 

\section{Ethical Considerations}

We collect the names of users in the human evaluation game's database simply for logistical purposes. Other than this, no personal data is collected. The data collected and its purpose were verbally conveyed to each evaluator before asking for their consent. We remove the names of evaluators in the data release. Besides this, we ensure that now and in the future, any data collection is transparent with users and is used in an ethical and responsible manner. Since our research primarily evaluates reasoning in a game environment, there are fewer potential real-world risks of its applications. However, biases in LLMs may be reproduced. 

% Bibliography entries for the entire Anthology, followed by custom entries
%\bibliography{anthology,custom}
% Custom bibliography entries only
\bibliography{acl_latex}

\appendix

\section{Prompt} \label{appendix:prompt}
Solve today’s NYT Connections game. 
Here are the instructions for how to play this game:\\
Find groups of four items that share something in common.\\
\textbf{Category Examples}:\\
FISH: Bass, Flounder, Salmon, Trout\\
FIRE \_\_\_:
Ant, Drill, Island, Opal\\
Categories will always be more specific than `5-LETTER-WORDS', `NAMES', or `VERBS.'
\\\\
\textbf{Example 1:}\\
\emph{Words}: [`DART', `HEM', `PLEAT', `SEAM', `CAN', `CURE', `DRY', `FREEZE', `BITE', `EDGE', `PUNCH', `SPICE', `CONDO', `HAW', `HERO', `LOO']\\
\emph{Groupings}:
\begin{enumerate}
\item Things to sew: [`DART', `HEM', `PLEAT', `SEAM’]
\item Ways to preserve food: [`CAN', `CURE', `DRY', `FREEZE’]
\item Sharp quality: [`BITE', `EDGE', `PUNCH', `SPICE’]
\item Birds minus last letter: [`CONDO', `HAW', `HERO', `LOO’]
\end{enumerate}
\textbf{Example 2:}\\
\emph{Words}: [1COLLECTIVE', `COMMON', `JOINT', `MUTUAL', `CLEAR', `DRAIN', `EMPTY', `FLUSH', `CIGARETTE', `PENCIL', `TICKET', `TOE', `AMERICAN', `FEVER', `LUCID', `PIPE']\\
\emph{Groupings}:
\begin{enumerate}
\item Shared: [`COLLECTIVE', `COMMON', `JOINT', `MUTUAL’]
\item Rid of contents: [`CLEAR', `DRAIN', `EMPTY', `FLUSH’]
\item Associated with “stub”: [`CIGARETTE', `PENCIL', `TICKET', `TOE’]
\item \_\_ Dream: [`AMERICAN', `FEVER', `LUCID', `PIPE’])
\end{enumerate}
\textbf{Example 3:}\\
\emph{Words}: [`HANGAR', `RUNWAY', `TARMAC', `TERMINAL', `ACTION', `CLAIM', `COMPLAINT', `LAWSUIT', `BEANBAG', `CLUB', `RING', `TORCH', `FOXGLOVE', `GUMSHOE', `TURNCOAT', `WINDSOCK']\\
\emph{Groupings}: 
\begin{enumerate}
\item Parts of an airport: [`HANGAR', `RUNWAY', `TARMAC', `TERMINAL’]
\item Legal terms: [`ACTION', `CLAIM', `COMPLAINT', `LAWSUIT’]
\item Things a juggler juggles: [`BEANBAG', `CLUB', `RING', `TORCH’]
\item Words ending in clothing: [`FOXGLOVE', `GUMSHOE', `TURNCOAT', `WINDSOCK’]
\end{enumerate}
Categories share commonalities:
\begin{itemize}
    \item There are 4 categories of 4 words each
    \item Every word will be in only 1 category
    \item One word will never be in two categories
    \item As the category number increases, the connections between the words and their category become more obscure. Category 1 is the most easy and intuitive and Category 4 is the hardest
    \item There may be a red herrings (words that seems to belong together but actually are in separate categories)
    \item Category 4 often contains compound words with a common prefix or suffix word
    \item A few other common categories include word and letter patterns, pop culture clues (such as music and movie titles) and fill-in-the-blank phrases
\end{itemize}
You will be given a new example (Example 4) with today’s list of words. First explain your reason for each category and then give your final answer following the structure below (Replace Category 1, 2, 3, 4 with their names instead)\\\\
Groupings: \\
Category1: [word1, word2, word3, word4] \\
Category2: [word5, word6, word7, word8] \\
Category3: [word9, word10, word11, word12] \\
Category4: [word13, word14, word15, word16] \\\\
Remember that the same word cannot be repeated across multiple categories, and you need to output 4 categories with 4 distinct words each. Also do not make up words not in the list. This is the most important rule. Please obey\\\\
\textbf{Example 4:}\\
Words : [InsertGame] \\
Groupings \\

\section{Disagreements in Annotations} \label{appendix:disagree}
There was some disagreement between Semantic and Encyclopedic Knowledge labeling. One example is the grouping \emph{pike, split, straddle, tuck} which are ``Gymnastics Positions" and requires domain-specific knowledge, meaning it could be thought of as Encyclopedic. However, it could also be classified under Semantic Relations (\textit{Type Of} relation) as many of these words appear in WordNet. There can be confusion about what is considered to be Semantic or Associative Knowledge. For example, \emph{card, chocolate, heart, rose} might be thought of to be semantically related as gifts given to a loved one and hence labeled under semantic relations; however, they can be viewed as elements of specific situations or thematic context such as ``Seen on Valentine's Day". Hence it falls more appropriately under Associative Relations.

\section{Red Herrings}\label{appendix:redherrings}
\begin{figure}[!ht]
    \includegraphics[scale=0.11]{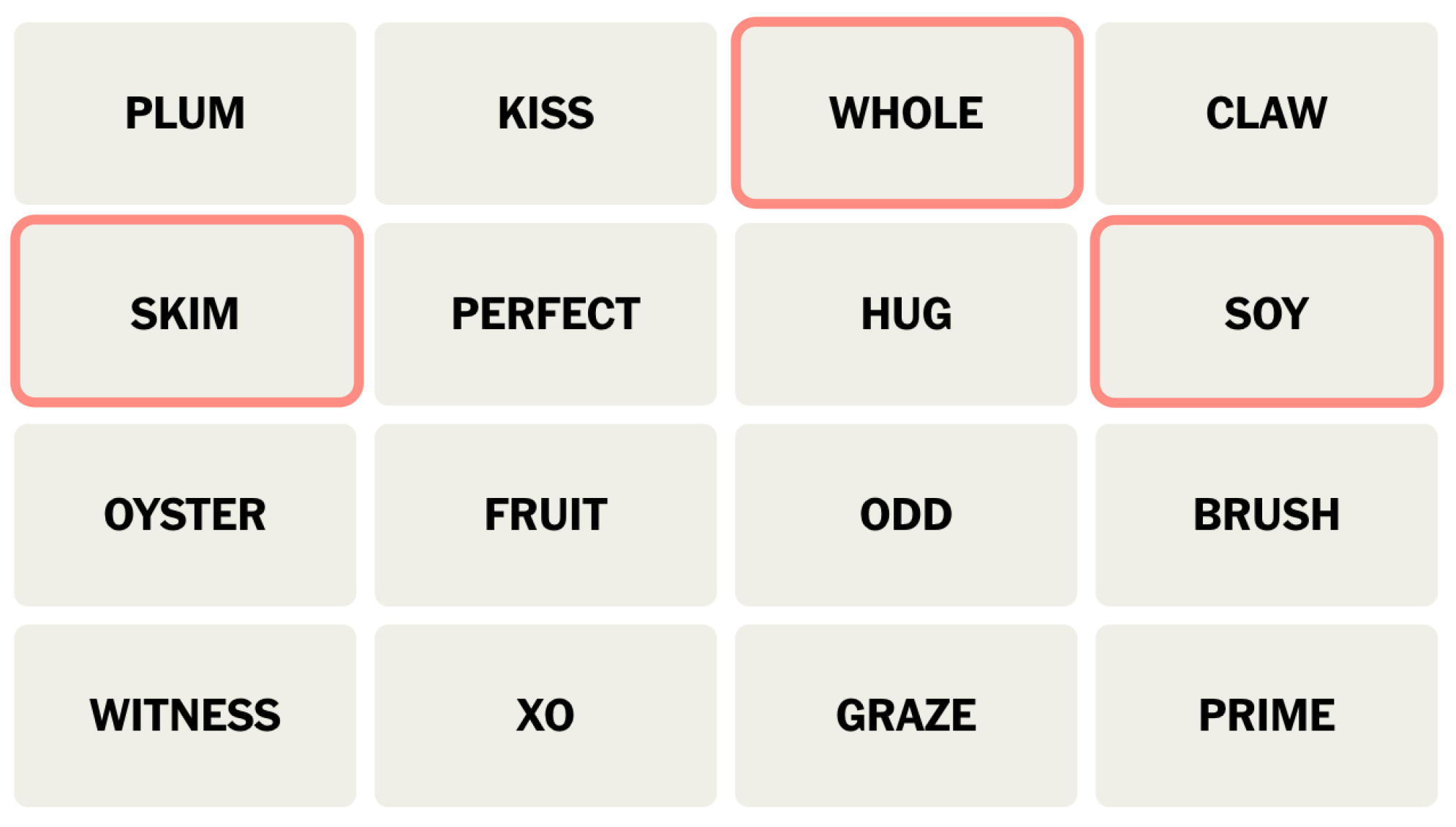}
    \caption{Example of red herring category where the 3 words outlined in red might seem as though they belong together.}
    \label{fig:redherringcategory}
\end{figure}
In the puzzle in Figure~\ref{fig:redherringcategory}, a red herring category is present. Gemini 1.5 Pro created a category called "Milk" with \emph{Whole, Skim,} and \emph{Soy} and included a random fourth word that did not fit. Each of these three words, however, belongs to a different category: \emph{Whole} to \emph{Kinds of Numbers}, \emph{Skim} to \emph{Touch Lightly}, and \emph{Soy} to \emph{Sauces in Chinese Cuisine}. In other puzzles including a red herring category like this one, most models make similar rationalizations.

\begin{figure}[!ht]
    \includegraphics[scale=0.2]{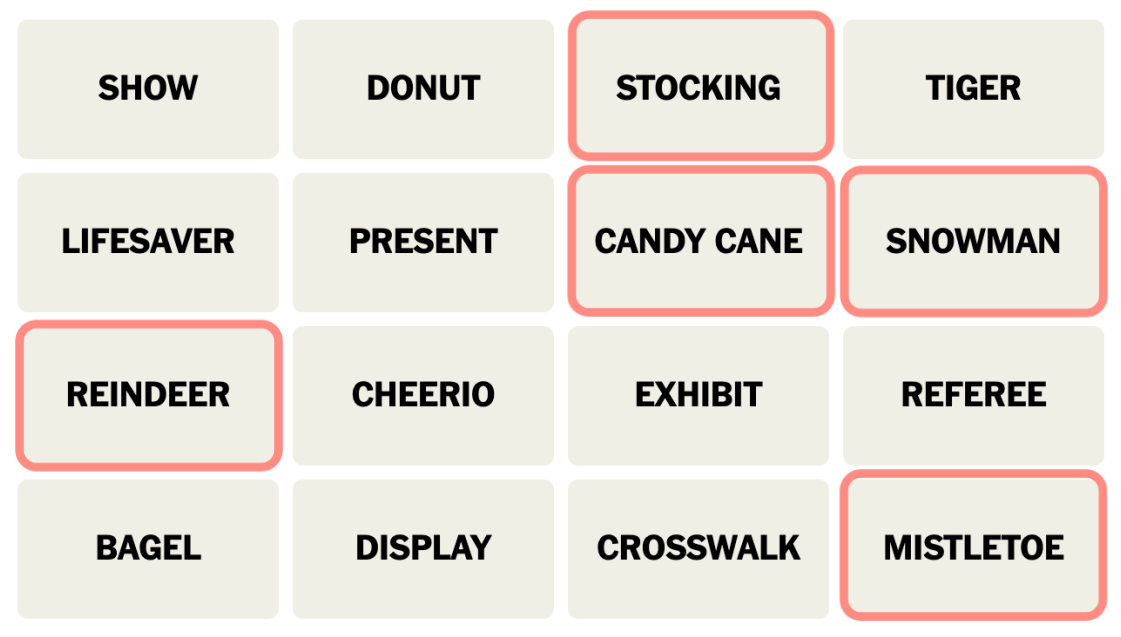}
    \caption{Example of red herring word where the 5 words outlined in red may seem like they belong together.}
    \label{fig:redherringword}
\end{figure}

The game in Figure~\ref{fig:redherringword} is an example of a game with a red herring word. The five words that appear as though they belong together are outlined in red. However, \emph{Mistletoe, Reindeer, Snowman,} and \emph{Stocking} form the ``Christmas Related" category, while \emph{Candy Cane} belongs to the category ``Things with Stripes". In this game, all models except Claude 3.5 Sonnet made the mistake of grouping \emph{Candy Cane} with some combination of three of the other Christmas-related words.

\section{Performance}\label{appendix:performance}
\subsection{Humans}\label{appendix:human_performance}
The frequency of clustering scores for novice human players in 100 games and scores for Claude 3.5 Sonnet in the same games are shown in Table~\ref{table: noviceclassification}. The frequency of clustering scores for expert human players in 50 games and scores for Claude 3.5 Sonnet in the same games are in Table~\ref{table: expertclassification}. 

\begin{figure}[!ht]
    \small
    \centering
    \includegraphics[scale=0.34]{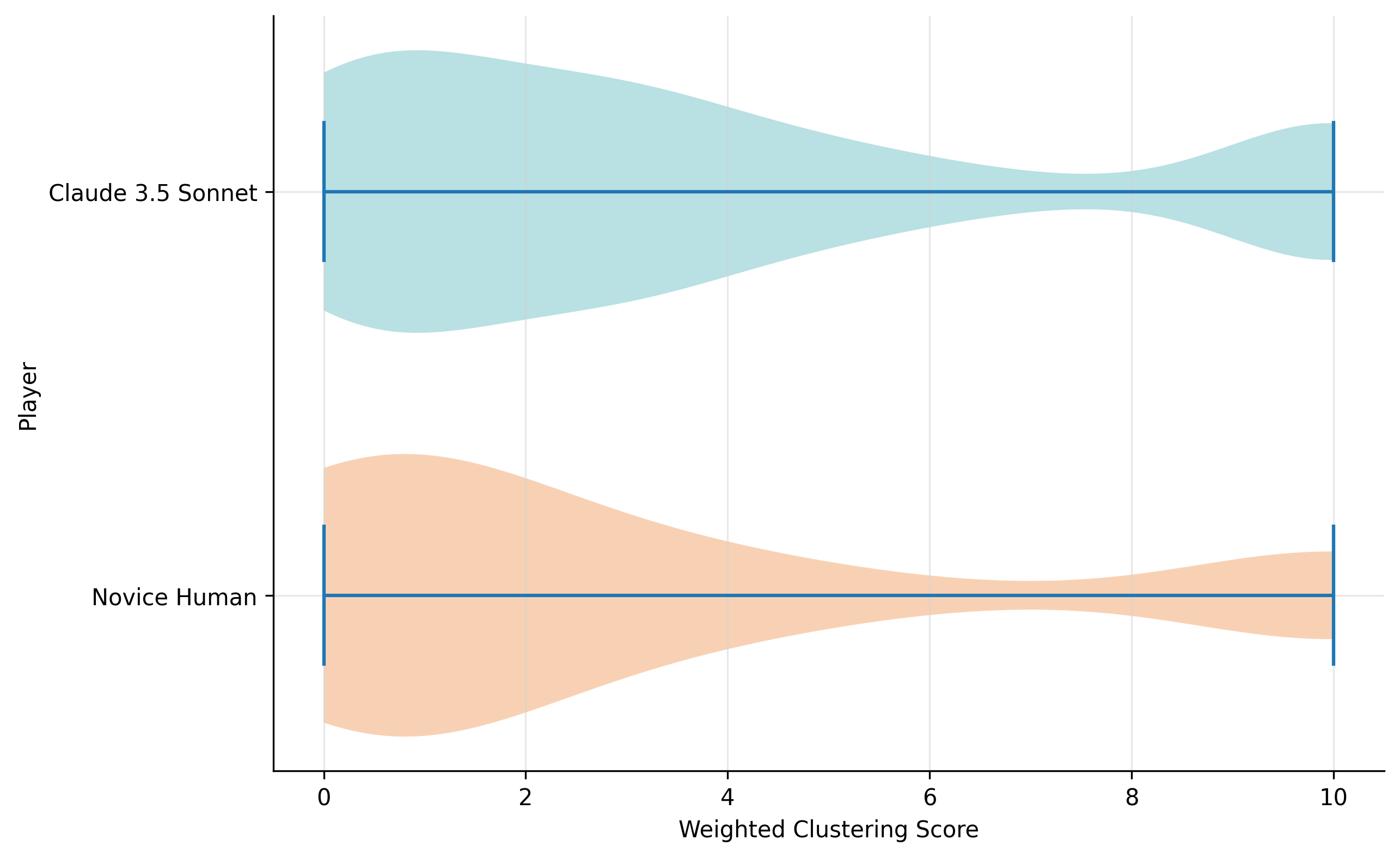}
    \caption{Spread of weighted clustering score for Claude 3.5 Sonnet and novice human players across 100\emph{NYT} Connections games}
    \label{fig:novice_weighted}
\end{figure}

\begin{figure}[!ht]
    \small
    \centering
    \includegraphics[scale=0.34]{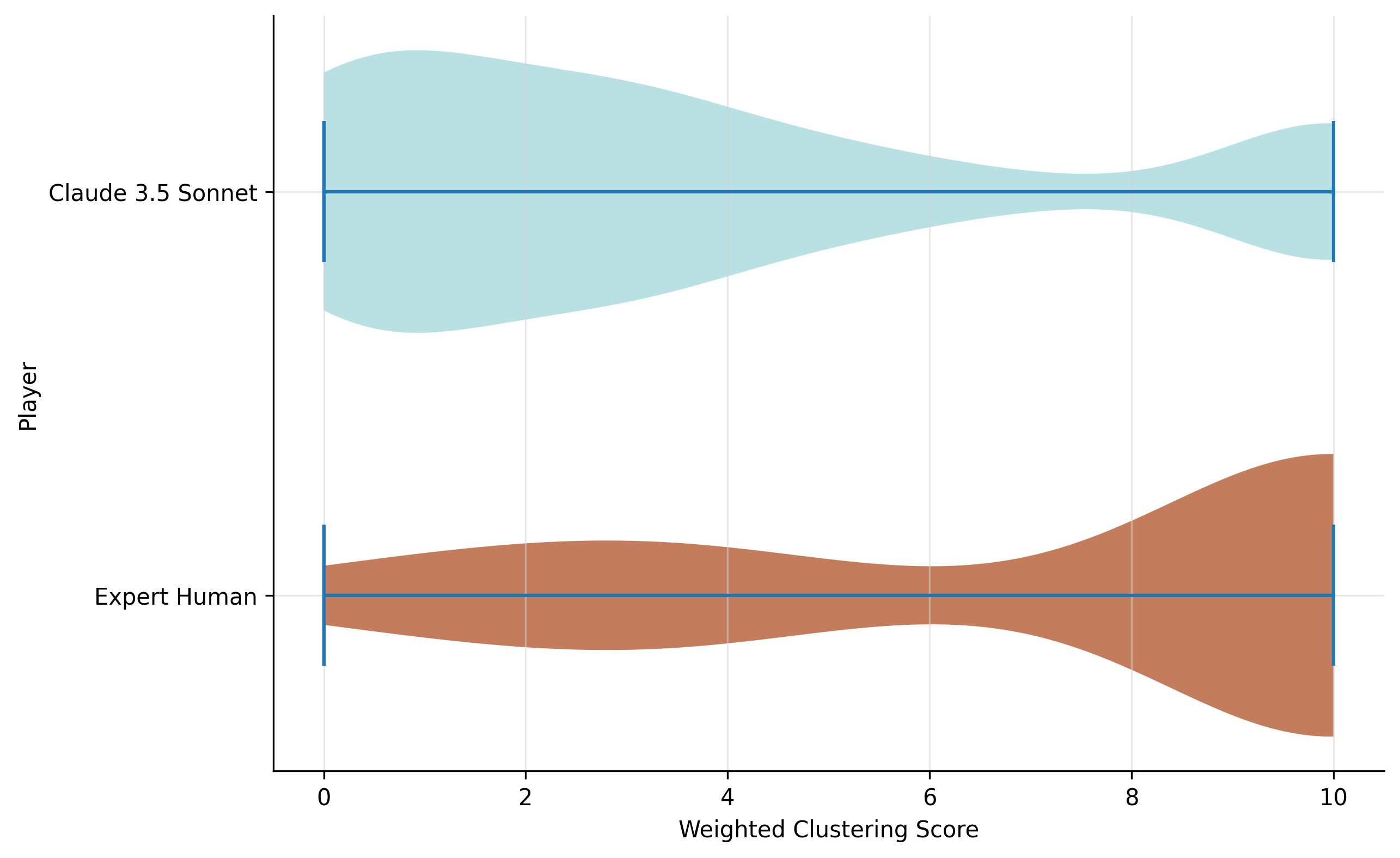}
    \caption{Spread of weighted clustering score for Claude 3.5 Sonnet and expert human players across 50\emph{NYT} Connections games}
    \label{fig:expert_weighted}
\end{figure}

Figures~\ref{fig:novice_weighted} and \ref{fig:expert_weighted} show the distribution of the weighted clustering scores for Claude 3.5 Sonnet against novice humans and expert humans, respectively. 

\begin{table}[!ht]
\centering
\begin{tabular}{|c|c|c|}
\hline
\begin{tabular}[c]{@{}l@{}}\textbf{Unweighted} \\ \textbf{Clustering}\\\textbf{Score}\end{tabular} & \begin{tabular}[c]{@{}l@{}}Claude 3.5 \\ Sonnet\end{tabular}& Novice Humans \\ \hline
0  & 17 & 30 \\ \hline
1  & 41 & 39  \\ \hline
2  & 27 & 13\\ \hline
3  & 3 & 0 \\ \hline
4  & 12 & 18\\ \hline
\end{tabular}
\caption{\label{table: noviceclassification}Frequency of clustering scores 0-4 for GPT-4o and novice human players across 100\emph{NYT} Connections games}
\end{table}

\begin{table}[!ht]
\centering
\begin{tabular}{|c|c|c|}
\hline
\begin{tabular}[c]{@{}l@{}}\textbf{Unweighted} \\ \textbf{Clustering}\\\textbf{Score}\end{tabular} & \begin{tabular}[c]{@{}l@{}}Claude 3.5 \\ Sonnet\end{tabular}& Expert Humans \\ \hline
0  & 6& 2 \\ \hline
1  & 20& 7 \\ \hline
2  & 13& 9 \\ \hline
3  & 1& 0 \\ \hline
4  & 10& 32 \\ \hline
\end{tabular}
\caption{\label{table: expertclassification}Frequency of clustering scores 0-4 for GPT-4o and expert human players across 50\emph{NYT} Connections games}
\end{table}

\begin{table*}[!ht]
\centering
\begin{tabular}{|c|c|c|c|c|c|}
\hline
\begin{tabular}[c]{@{}l@{}}\textbf{Unweighted} \\ \textbf{Clustering}\\\textbf{Score}\end{tabular} & Gemini 1.5 Pro & Claude 3.5 Sonnet & GPT-4o & Llama 3.1 405B & Mistral Large 2 \\ \hline
0  & 161 & 87 & 112 & 146 & 185\\ \hline
1  & 153 & 138 & 140 & 150 & 149\\ \hline
2  & 90 & 117 & 111 & 93 & 80\\ \hline
3  & 10 & 17 & 10 & 2 & 3\\ \hline
4  & 24 & 79 & 65 & 47 & 21\\ \hline
\end{tabular}
\caption{\label{table:llmclassification}Frequency of unweighted clustering scores 0-4 for 5 LLMs across 438\emph{NYT} Connections games}
\vspace{-0.7em}
\end{table*}

\begin{table*}[!ht]
\centering
\begin{tabular}{|c|c|c|c|c|c|}
\hline
\begin{tabular}[c]{@{}l@{}} \textbf{Categorical} \\ \textbf{Reasoning}\\\textbf{Score} \end{tabular}& Gemini 1.5 Pro & Claude 3.5 Sonnet & GPT-4o & Llama 3.1 405B & Mistral Large 2 \\ \hline
0 & 203 & 101 & 131 & 176 & 215\\ \hline
1 & 154 & 152 & 157 & 151 & 144\\ \hline
2 & 59 & 109 & 94 & 73 & 61\\ \hline
3 & 16 & 53 & 36 & 25 & 13\\ \hline
4 & 6 & 23 & 20 & 13 & 5\\ \hline
\end{tabular}
\caption{\label{table:categorical}Frequency of categorical reasoning scores 0-4 for 5 LLMs across 438\emph{NYT} Connections games}
\end{table*}

\subsection{LLMs}\label{appendix:llm_performance}
Table~\ref{table:llmclassification} shows the frequency of the unweighted clustering scores (number of categories correctly grouped) for each LLM. The total number of games played by each model is 438. Table~\ref{table:categorical} is slightly different and shows the frequency of categorical reasoning scores (the categories correctly grouped and reasoned) for each model. Because a caveat for receiving a categorical reasoning score greater than 0 is matching gold category words and names, a score of 0 is more common than in the unweighted clustering scores.

\section{Human Evaluation Interface} \label{sec:humanevalinterface}
\begin{figure*}[!ht]
    \centering
    \subfloat[Instruction screen]{\includegraphics[scale=0.3]{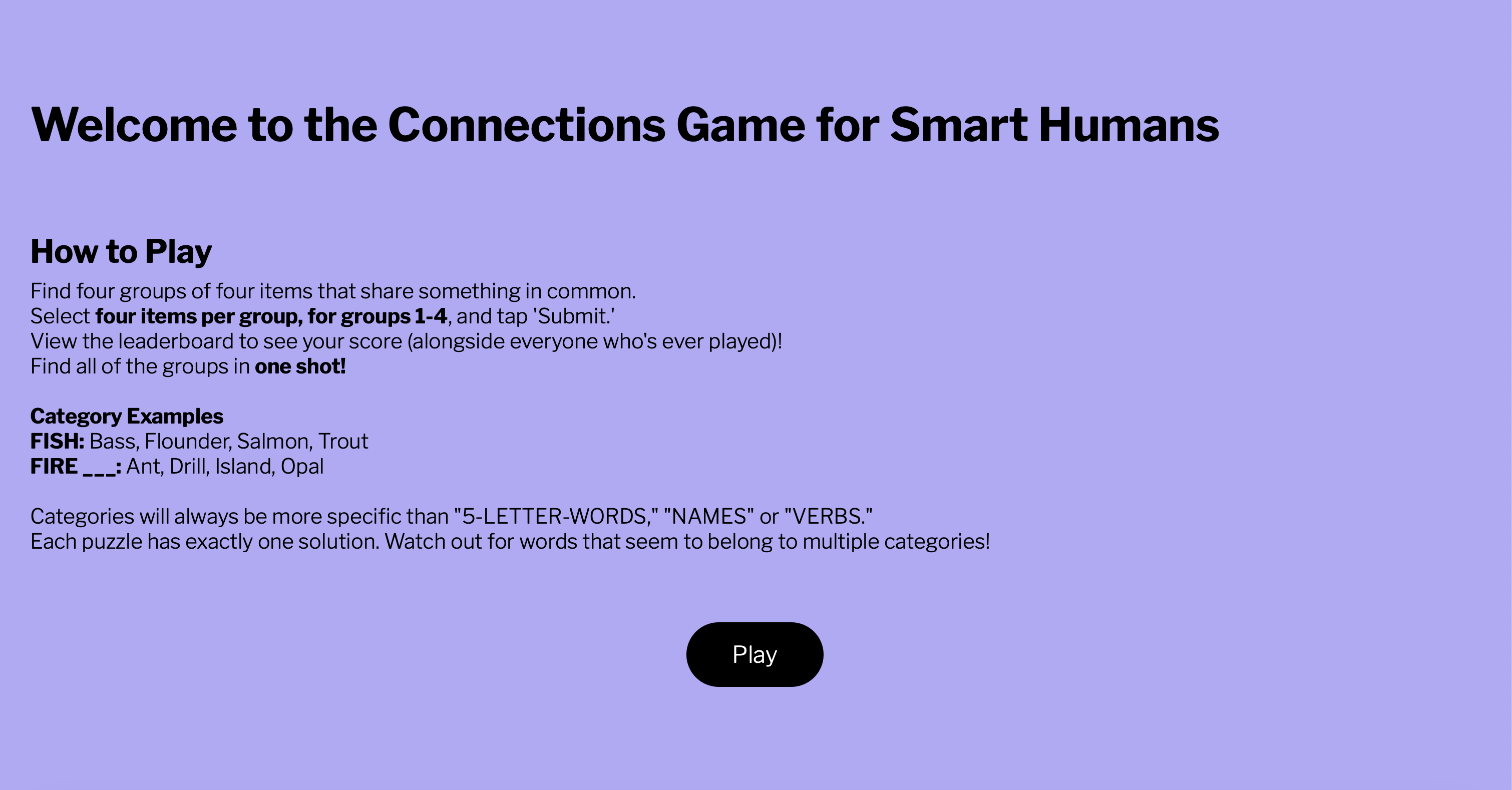}}\\
    \subfloat[Example of game play]{\includegraphics[scale=0.48]{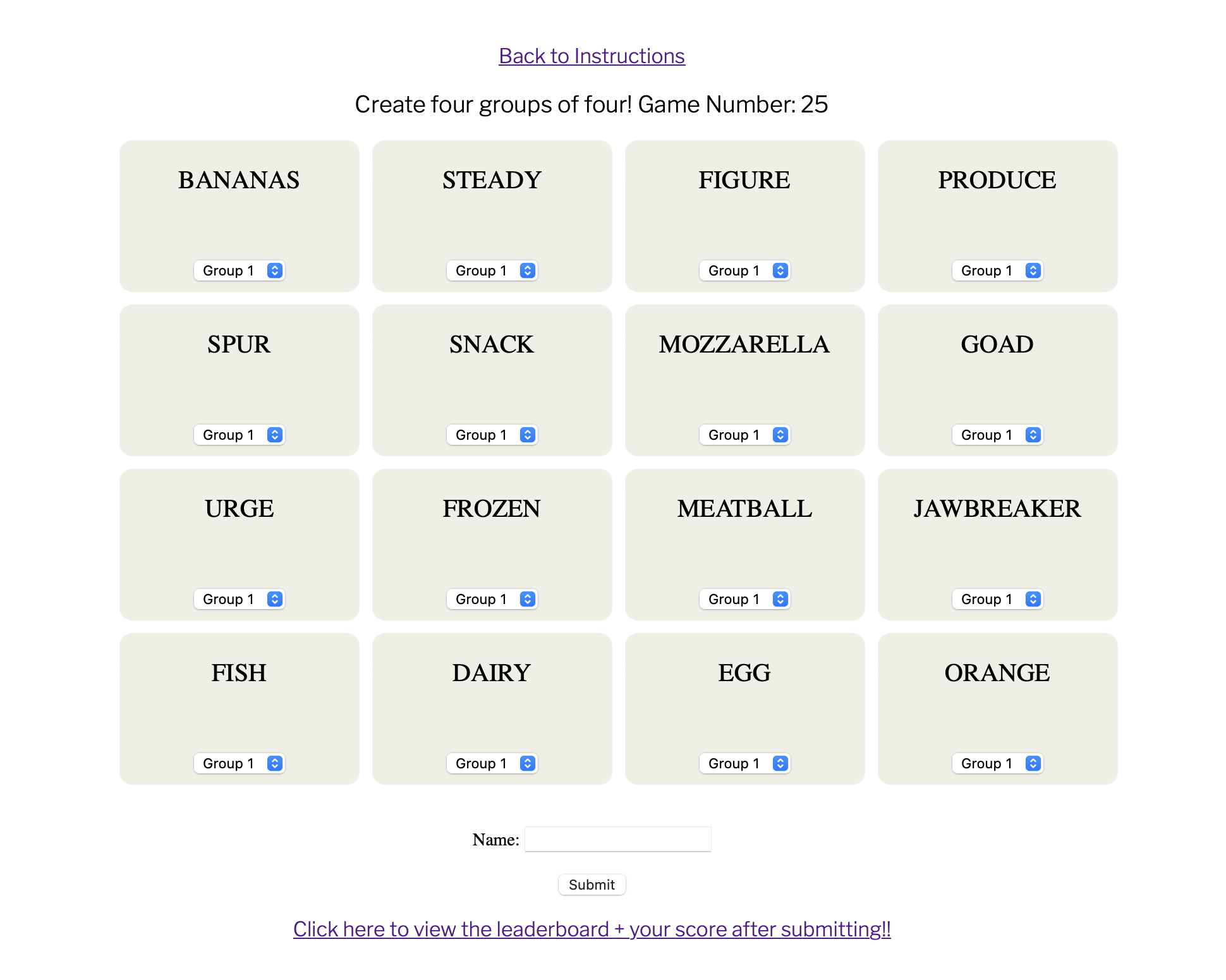}}
    \caption{Human evaluation interface}
    \label{fig:human_interface}
\end{figure*}

Figure~\ref{fig:human_interface} shows the two main screens of the evaluation interface provided to both novice and expert human players. (a) is the instruction screen, while (b) is an example of a game screen after the user hits the "Play" button. To solve the game in one shot, all 16 words from a game are displayed on the screen in separate boxes, with one drop-down per box. The drop-down consists of four labels: Group 1, Group 2, Group 3, and Group 4. The user's job is to create 4 groups of 4 words using the given labels. Because the groups are chosen from a drop-down menu where the default option is Group 1, a clustering score of 3 is impossible. 

We stored the data collected in a SQLite database. Other than any name of choice users were prompted to enter in the "Name" text entry box, no personal data was collected. Each evaluator was then assigned initials in the final dataset collected for evaluation. These initials are not included in the data release. The data that would be collected and its purpose were verbally conveyed to each evaluator before asking for their consent. 
%  \begin{figure}[*ht]
%     \centering
%     \includegraphics[scale=0.25]{pictures/erd.png}
%     \caption{Human evaluation interface database schema}
%     \label{fig:sqlschema}
% \end{figure}

\end{document}